\documentclass[12pt,reqno]{article}
\usepackage{graphicx}
\usepackage{amsmath}
\usepackage{times}
\usepackage{hyperref}
\usepackage{float}
\usepackage{accsupp}
\usepackage{fullpage}
\usepackage{subcaption} 
\usepackage{caption}
\usepackage{longfigure}
\usepackage{booktabs}
\usepackage{caption}
\usepackage{amsmath}
\usepackage{amsfonts}   
\usepackage{enumitem}

\begin{document}

\title{
Correction of Transformer-Based Models with Smoothing Pseudo-Projector 
}


\author{
Vitaly Bulgakov$^{1,2}$ \\
\small
$^{1}$Profiteya LLC, Boston, MA, USA \\
\small
$^{2}$Mass General Brigham, Boston, MA, USA
}

\date{March 10, 2026}

\maketitle

\begin{abstract}
The pseudo-projector is a lightweight modification that can be integrated into existing language models and other neural networks without altering their core architecture. It can be viewed as a hidden-representation corrector that reduces sensitivity to noise by suppressing directions induced by label-irrelevant input content.
The design is inspired by the multigrid (MG) paradigm, originally developed to accelerate the convergence of iterative solvers for partial differential equations and boundary value problems, and later extended to more general linear systems through algebraic multigrid methods.
We refer to the method as a pseudo-projector because its linear prototype corresponds to a strictly idempotent orthogonal projector, whereas the practical formulation employs learnable restriction and prolongation operators and therefore does not, in general, satisfy the properties of an exact orthogonal projection.
We evaluate the proposed approach on transformer-based text classification tasks, as well as controlled synthetic benchmarks, demonstrating its effectiveness in improving training dynamics and robustness.
Experimental results, together with supporting theoretical heuristics, indicate consistent improvements in training behavior across a range of settings, with no adverse effects observed otherwise. Our next step will be to extend this approach to language models. 
\end{abstract}

\section{Introduction}
This work is motivated by prior research on algebraic multigrid preconditioners for complex structural systems conducted by the author, as well as by ongoing efforts to transfer multigrid ideas to neural network optimization.

\emph{The purpose of this study is to introduce a lightweight enhancement that can be incorporated into complex models without disrupting their core architectural components, such as attention mechanisms, transformer layers, and feed-forward blocks}. It is also not intended to operate at the level of loss optimization, but rather to provide an enhancement to the training dynamics within the model design. 

A fundamental challenge in training neural networks is the highly non-convex nature of the optimization landscape, which can lead to slow convergence or stagnation in suboptimal local minima or saddle regions, hindering progress toward globally optimal solutions.
We propose a pseudo-projector that, in the case of linear systems, reduces to an orthogonal projector defined by prolongation and restriction operators. When applied to a hidden layer, this operator provides residual smoothing that may, in many cases, lead to substantial improvements in model training, while not adversely affecting performance in other cases.

We conducted two sets of experiments on binary classification tasks. The first experiment considers a synthetic dataset defined by a “wiggly” decision boundary that combines a globally convex trend with superimposed oscillatory components, together with a cloud of two-dimensional (x,y) points distributed above and below the curve. Using a simple neural network augmented with the proposed smoothing pseudo-projector, we graphically demonstrate how the inclusion of the projector alters the training dynamics and improves the learned decision boundary. 
The second set of experiments was conducted on real-world text classification datasets using a simple transformer-based model. The results show that incorporating the smoothing pseudo-projector can substantially accelerate the training process and, in some cases, enable the model to achieve higher evaluation metrics than the same architecture trained without the projector. To more rigorously evaluate the benefits of the smoothing corrector, we deliberately introduce additional challenges by violating class balance and injecting noisy content into both training and validation samples.

\section{Related Work}
Early work on multigrid (MG) methods focused on accelerating the convergence of iterative solvers for systems of equations arising from the discretization of boundary value problems. The first explicit two-grid algorithm was proposed by Fedorenko [1]. Brandt subsequently established multigrid as a systematic numerical method by introducing key concepts such as V- and W-cycles, smoothing procedures, and coarse-grid correction [2].
Subsequent efforts sought to extend multigrid ideas to problems involving complex grids or to settings independent of geometric information, leading to the development of Algebraic Multigrid (AMG). These methods apply multigrid principles directly to matrix equations without relying on an explicit grid structure. Representative works in this direction, largely of a theoretical nature, include those by Ruge and Stüben [3], Hackbusch [4], and Brandt [5].

Multilevel aggregation methods for solving complex structural analysis problems were proposed in [6, 7] and further developed in follow-up studies. These methods are based on iterative solvers equipped with preconditioners that incorporate a coarse model constructed through aggregation of complex structures represented using finite element approximations.
There have also been works extending multigrid methods to nonlinear problems via the Full Approximation Scheme (FAS), introduced by Brandt [2], which enables coarse-grid correction for nonlinear systems. Further research established both theoretical foundations and practical algorithms for nonlinear multigrid and multigrid-based optimization methods (e.g., [8, 9]).

With the advent of massive applications of deep neural networks and the growing recognition of optimization challenges associated with highly non-convex loss landscapes, multigrid (MG) ideas have increasingly been explored in the context of neural network training. Applications of multigrid ideas to neural networks can be broadly divided into two categories. The first applies multigrid principles at the level of loss optimization, modifying the training dynamics through coarse-to-fine optimization, multigrid-in-time, or preconditioning techniques while keeping the network architecture fixed. The second embeds multigrid concepts within the network architecture or hidden representations, introducing multiscale, projection, or smoothing operators that alter information flow without modifying the loss function or optimization algorithm. Some methods exhibit characteristics of both categories. 

Examples of such approaches include the following. In [10], a parallel nonlinear multigrid iteration is applied across network layers by viewing depth as a temporal dimension, replacing sequential forward and backward passes with multigrid iterations and allowing inexact gradient information; as a result, the training dynamics are altered through a multigrid solve. Work [11] explicitly applies multigrid reduction in time (MGRIT) to compute hidden states for both forward and backward propagation, directly affecting how gradients are generated and propagated. Similarly, [12] formulates layer-wise propagation as a coupled system and employs MGRIT-style iterations to enable scalable training. In [13], convolutional neural networks are extended to operate across multiple scales or grids simultaneously, allowing features at different resolutions to interact in a manner analogous to multigrid feature spaces rather than classical PDE (Partial Differential Equations) solvers. Reference [14] applies a multigrid decomposition across channel hierarchies in CNNs to reduce parameter redundancy, effectively embedding multigrid structure into the network representation. The authors of [15] draw mathematical and structural analogies between multigrid methods used for solving PDEs and architectural components of CNNs, such as pooling operations and iterative feature extraction. [16] employs multiresolution and multigraph decompositions to incorporate multigrid principles into neural operators for PDE learning. [17] demonstrates algebraic multigrid using graph neural networks with learned MG operations. This list of works is certainly not exhaustive and includes only a selection of relevant studies.

\section{Smoothing Pseudo-Projector}

We want to emphasize from the very beginning that the discussion of the intended properties of the projector that follows is based on assumptions and primarily motivated by intuition, heuristic arguments, and empirical observations, rather than by formal mathematical proofs, as often the case in description of complex methods of applied mathematics. The mathematical arguments are heuristic and are included to support intuition and interpretation of the experimental findings.

We refer to the operator as a pseudo-projector because, while it reduces to an orthogonal projector in the linear case, its trainable parameters in the neural network setting generally break exact orthogonality, resulting in an approximate projection that acts as a residual smoothing operator.

The linear prototype of the proposed pseudo-projector is the standard orthogonal
projection operator constructed from prolongation and restriction operators.
Let \( Q : \mathbb{R}^c \to \mathbb{R}^f \) denote a prolongation operator mapping
a coarse space of dimension \( c \) to a fine space of dimension \( f \), and let
\( Q^{*} : \mathbb{R}^f \to \mathbb{R}^c \) denote the corresponding restriction
operator. Here, the notation \( c \) and \( f \) reflects the classical multigrid
distinction between coarse and fine levels, with \( c \ll f \).
The associated projection operator is given by
\[
P = Q \left( Q^{*} Q \right)^{-1} Q^{*}.
\]

Assuming that \( Q \) has full column rank, \( P \) is the orthogonal projector
onto the range of \( Q \). In a Euclidean setting, \( P \in \mathbb{R}^{f \times f} \)
is symmetric and idempotent, with rank equal to \( c \), corresponding to the
dimension of the coarse subspace embedded in the fine space.

In the neural network setting, the projection operator \( P \) is applied to
intermediate hidden representations in order to introduce multigrid-inspired
residual smoothing without altering the core architecture. Let
\( h \in \mathbb{R}^f \) denote a hidden representation produced by a network
layer (e.g., the output of an attention or feed-forward block).
The pseudo-projector is applied in a residual manner,
\begin{equation}
h \leftarrow h + P h \label{eq:proj_interpolation}
\end{equation}
In this context, the range of Q represents a low-dimensional coarse subspace embedded in the full hidden space, intended to capture dominant, low-frequency or global components of the representation. \emph{By projecting h onto this subspace and re-injecting the result into the residual stream, the operator selectively smooths high-frequency components while preserving the original representation}. This mechanism improves the conditioning of the training dynamics and facilitates information propagation across layers, while leaving the loss function, optimization algorithm, and primary architectural components unchanged.

In transformer-based language models, the hidden-layer representation \( h \)
is multi-dimensional. A typical tensor shape is \( (B, T, D) \), where
\( B \) denotes the batch size, \( T \) the sequence length, and
\( D \) the dimensionality of the token representation (embedding size).
In this work, the projector operates on the activation (feature) dimension D, the sequence (temporal) dimension T, or on both dimensions simultaneously. 

If we introduce parameter \(\alpha\), (1) can be rewritten in the form
\begin{equation}
h' = M h, \qquad
M := P + \alpha (I - P), \qquad
\alpha \in [0,1],
\label{eq:proj_interpolation}
\end{equation}
which induces an orthogonal decomposition of the hidden representation \( h \)
into coarse and complementary components.
Specifically, the projector \( P \) extracts the component of \( h \) lying in
the coarse subspace, while \( I - P \) projects onto its orthogonal complement.
The parameter \( \alpha \) controls the relative contribution of the
complementary component, interpolating between full projection
(\( \alpha = 0 \)) and the identity mapping (\( \alpha = 1 \)).

As a result, the transformation M selectively amplifies coarse-scale structure while damping high-frequency components, providing a controllable residual smoothing mechanism within the network.

\section{Intuition and Mathematical Heuristics}

Let a model produce a hidden representation \( h(x) \in \mathbb{R}^D \) and let a
classifier head \( g \) produce logits
\[
z(x) = g(h(x)).
\]
\textbf{Assumption A1 (Signal lives in the coarse subspace).}
For inputs \( x \) and labels \( y \in \{+1,-1\} \), decompose
\[
h(x) = s_y(x) + n(x),
\]
with
\[
s_y(x) \in \operatorname{range}(P), \qquad
n(x) \in \operatorname{range}(I - P).
\]
Where s denotes Signal and n denotes Noise.
\\\\
\textbf{Assumption A2 (Complement is mostly noise).}
The noise component \( n(x) \) is zero-mean and (approximately) isotropic in the
complementary subspace:
\[
\mathbb{E}[n(x) \mid y] = 0, \qquad
\operatorname{Cov}(n(x) \mid y) = \sigma^2 I
\quad \text{on } \operatorname{range}(I - P).
\]

Consider a linear classifier head (binary classification)
\( z = w^{\top} h \). Decompose the weight vector as
\[
w = w_c + w_f, \qquad
w_c := P w, \qquad
w_f := (I - P) w.
\]
Then, after smoothing,
\[
z' = w^{\top} M h
   = w^{\top} \bigl( P s_y + \alpha (I - P) n \bigr)
   = w^{\top} P s_y + \alpha w^{\top} (I - P) n
   = w_c^{\top} s_y + \alpha w_f^{\top} n,
\]
where we used \( P s_y = s_y \).

\begin{itemize}
\item \textbf{The mean margin stays the same:}
\[
\mathbb{E}[y z' \mid y] = y\, w_c^{\top} s_y .
\]

\item \textbf{The noise variance shrinks by \( \alpha^2 \):}
\[
\operatorname{Var}(z' \mid y)
= \alpha^2 \operatorname{Var}(w_f^{\top} n \mid y)
= \alpha^2 \sigma^2 \lVert w_f \rVert^2 .
\]
\end{itemize}

If \( w_f^{\top} n \) is approximately Gaussian (as commonly justified by
central limit arguments in high dimensions), then the misclassification
probability satisfies
\[
\Pr(y z' \le 0)
\approx
\Phi\!\left(
-\frac{y\, w_c^{\top} s_y}{\alpha \sigma \lVert w_f \rVert}
\right),
\]
where \( \Phi \) denotes the standard normal cumulative distribution function.

So smaller \( \alpha \) increases the effective margin-to-noise ratio and reduces
error, hence can improve validation accuracy.
We can interpret it this way. The complement, \( I - P \) projection, is features
that are not in a signal subspace. They are those that the projector removes.
We assume that they have nuances—patterns that help fit training, but do not
generalize. So, shrinking them prevents the model from relying on unstable,
non-generalizable features \( \rightarrow \) less overfitting, better validation
performance. On the other hand, if some discriminative signal lies in
\( \operatorname{range}(I - P) \) then shrinking these components will also
suppress useful information, reducing class separability and hurting performance.
So \( \alpha \) controls the bias--variance tradeoff.

Another consideration is that small differences between samples in feature space
should not cause large changes in predictions. This is what stability and
generalization arguments care about. In this context let \( h_1, h_2 \in \mathbb{R}^f \)
denote hidden representations produced by the network for two input samples.
In case of a linear classifier with bounded weight norm or a multilayer perceptron
with bounded operator norms prediction head \( g \) can be considered
\( L_g \)-Lipschitz [18]. Then
\[
\lVert g(u) - g(v) \rVert \le L_g \lVert u - v \rVert
\quad \forall\, u, v .
\]
With the operator norm definition, we have:
\[
\lVert M x \rVert \le \lVert M \rVert_2 \, \lVert x \rVert
\quad \forall\, x .
\]
Then set \( u = M h_1 \), \( v = M h_2 \):
\[
\lVert g(M h_1) - g(M h_2) \rVert
\le L_g \lVert M h_1 - M h_2 \rVert
= L_g \lVert M (h_1 - h_2) \rVert
\le L_g \lVert M \rVert_2 \, \lVert h_1 - h_2 \rVert .
\]
For an orthogonal projector \( P \), the operator norm of \( M \) from (2) satisfies
\[
\lVert M \rVert_2 = \max(1, \alpha) = 1 .
\]
So, for any two samples with hidden representations h1 and h2, the smoothing operator contracts their separation in the complementary subspace by a factor \(\alpha<1\), while leaving coarse-scale components unchanged. As a result, the downstream predictor becomes less sensitive to sample-to-sample variability that does not contribute to the dominant structure of the data, which can lead to improved stability and generalization performance.

\section{Synthetic data experiment with “wiggly” decision boundary}

The purpose of this experiment is to isolate and visually demonstrate the effect
of the proposed pseudo-projector on training dynamics and decision boundary
formation in a controlled setting. By using a low-dimensional synthetic
classification task with a known, highly non-convex decision boundary, we can
directly assess how residual smoothing introduced by the projector influences
convergence, stability, and generalization, independent of architectural
complexity.

We construct a synthetic two-dimensional binary classification dataset in which
labels are determined by the position of points relative to a predefined
``wiggly'' nonlinear boundary. Each sample
\( x = (x_1, x_2) \in \mathbb{R}^2 \) is labeled as positive if \( x_2 \) lies above
the boundary and negative otherwise. The boundary function combines a smooth
global component with localized structure and high-frequency oscillations,
producing a decision surface that is challenging for small neural networks to
approximate.
Gaussian noise is added to the input features to introduce stochastic variability
and to better reflect realistic training conditions.

The pseudo-projector is implemented using a pair of learnable linear operators
corresponding to restriction (\( Q^{*} \)) and prolongation (\( Q \)), forming an
approximate orthogonal projection onto a lower-dimensional coarse subspace. At
each forward pass, the projection is computed by solving a small linear system
involving \( Q^{*} Q \) in the coarse space, with numerical stability ensured via
diagonal regularization. This operator is shared across layers and introduces
only a small number of additional parameters, making it a lightweight
architectural modification. With diagonal regularization, the projector has the form
\[
P = Q \bigl( Q^{*} Q + \varepsilon I \bigr)^{-1} Q^{*}.
\]
The pseudo-projector employs a restriction operator
\( Q^{*} : \mathbb{R}^D \to \mathbb{R}^{D_c} \) and a prolongation operator
\( Q : \mathbb{R}^{D_c} \to \mathbb{R}^D \), implemented as trainable, bias-free
linear layers \texttt{nn.Linear(D, D\_c)} and \texttt{nn.Linear(D\_c, D)} in terms of PyTorch objects assignment.
Concretely, these mappings are realized as
\( Q^{*}(h) = W_{Q^{*}} h \) with
\( W_{Q^{*}} \in \mathbb{R}^{D_c \times D} \) and
\( Q(z) = W_Q z \) with
\( W_Q \in \mathbb{R}^{D \times D_c} \), enabling learned coarse-to-fine
transformations that produce adaptive smoothing of hidden representations.

We employ a simple fully connected neural network (ToyNet) consisting of multiple
hidden layers with \texttt{tanh} activations and a linear output layer for binary
classification. To enable controlled experimentation, the same base architecture
is trained in two variants:
\begin{enumerate}
\item a plain model without projection, and
\item a projector-augmented model, in which the proposed pseudo-projector is
applied to each hidden representation.
\end{enumerate}
Hidden layers produce representations \( h \in \mathbb{R}^D \), to which the
projector is applied in residual form
\[
h \leftarrow \alpha h + (1 - \alpha) P(h),
\]
according to (2), where \( P \) is implemented via learnable restriction and prolongation operators
and \( \alpha \in (0,1) \) is a trainable scalar parameter. This formulation
preserves the original representation while selectively injecting a smoothed
coarse-scale component. Both models are trained under identical conditions using the Adam optimizer and cross-entropy loss.
Training and validation sets are generated independently to ensure unbiased evaluation. Performance is assessed using standard classification metrics (accuracy, precision, recall, and F1-score), as well as training and validation loss. In addition, decision boundaries learned by each model are visualized to qualitatively compare how the projector affects boundary smoothness and alignment with the ground-truth curve. Figure 1 shows the model with Projector diagram.
\begin{figure}[H]
\centering
\includegraphics[width=0.5\linewidth]{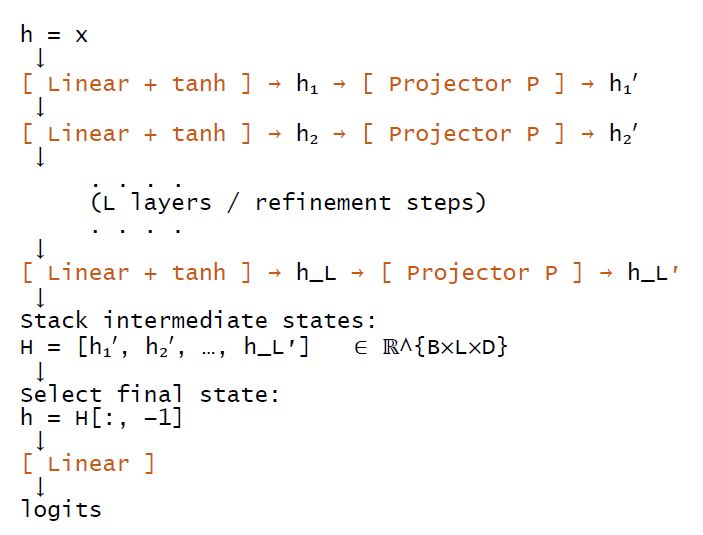}
\caption{Fig 1. “Wiggly curve" model structure}
\end{figure}

\subsection{Test 1}

In this experiment, training and validation were performed using the input parameters shown in Figure 2. The true decision boundary is globally convex and perturbed by high-frequency noise. The results depicted in Figure 3 illustrate that the projector improves the learned decision boundary by enhancing global shape adaptation and reducing sensitivity to local distortions. Figure 4 shows substantially faster convergence across all four validation metrics and for both training and validation losses. The same performance levels are achieved in significantly fewer epochs when the projector function is enabled.

\begin{figure}[t]
\centering
\includegraphics[width=0.6\linewidth]{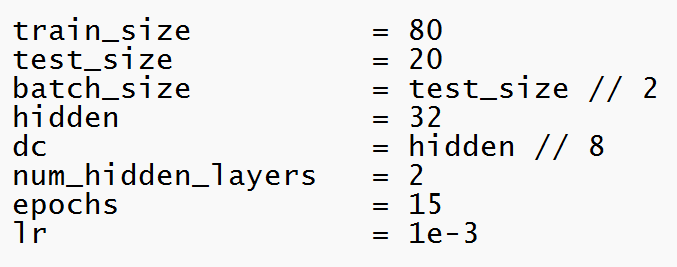}
\caption{Figure 2 Input parameters in test 1 including train and test sizes, batch size, hidden layers size, restriction (coarse) dimension, number of hidden layers, number of epochs and learning rate.}
\end{figure}

\begin{figure}[t]
\centering
\includegraphics[width=0.8\linewidth]{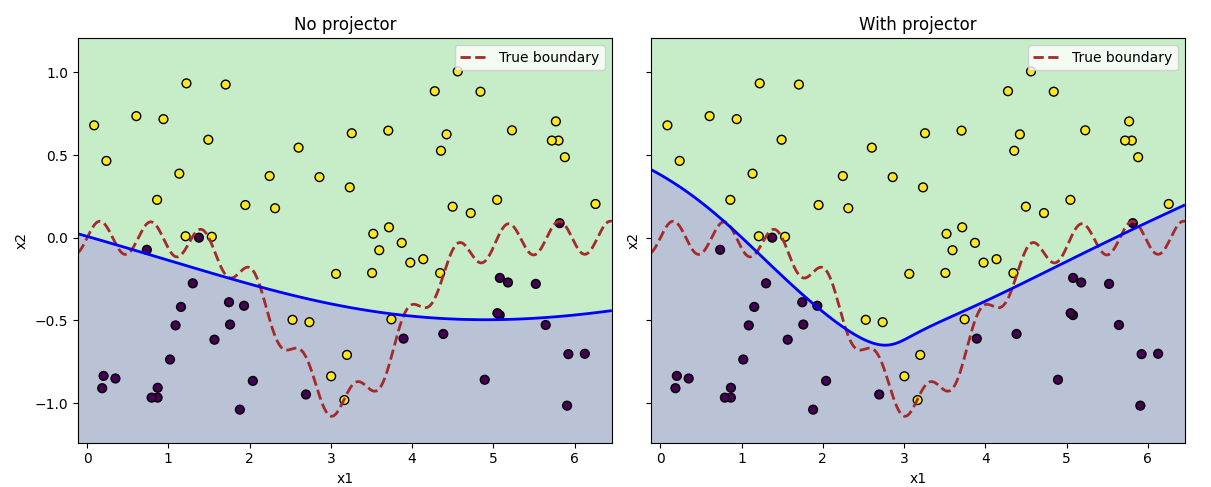}
\caption{Figure 3 compares the learned decision boundaries of the base model with and without the projector in test1.}
\end{figure}

\begin{figure}[t]
\centering
\includegraphics[width=0.8\linewidth]{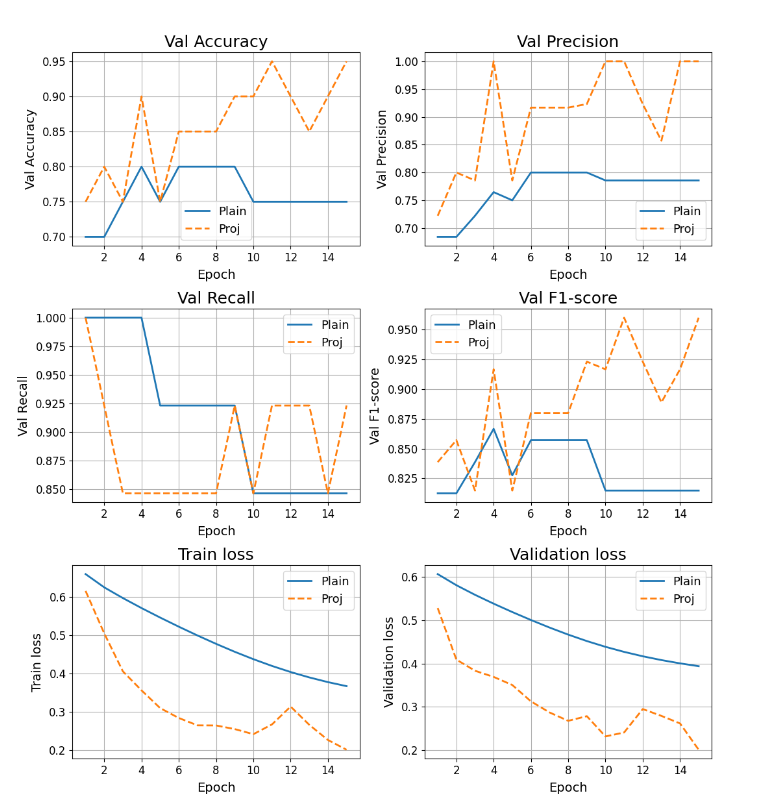}
\caption{Figure 4 Validation metrics and losses with training history in test1.}
\end{figure}

\subsection{Test 2}
In Test2, we increased the number of samples to 800 for training and 200, and demonstrate the obtained decision boundary shown in Figure 5 after the same 15 training epochs. As we can see from this plot the learned decision boundary again much better fits the original boundary in case of using correction projector.

\subsection{Test 3}
In Test 3, we further increased the number of samples to 8000 for training and 2000 for testing, and obtained the decision boundary plots shown in Figure 6 after the same 15 training epochs and in Figure 7 after 40 epochs.

\begin{figure}[H]
\centering
\includegraphics[width=0.7\linewidth]{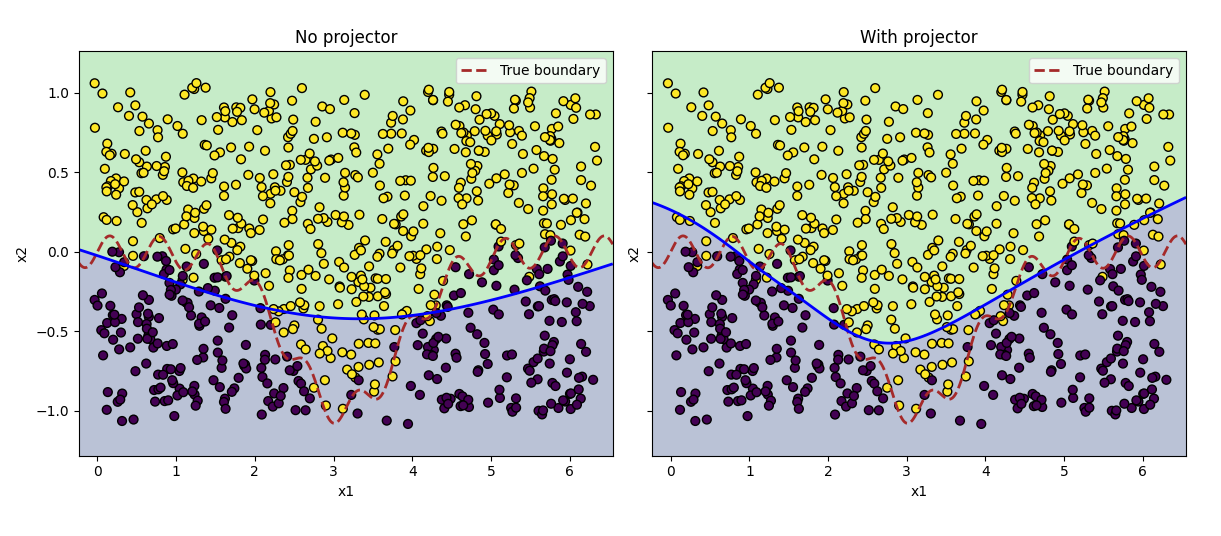}
\caption{Figure 5 compares the learned decision boundaries of the base model with and without the projector in test2 with 800 and 200 training and test samples accordingly after 15 epochs in Test 2, where solid blue line shows learned boundaries.}
\end{figure}

\begin{figure}[H]
\centering
\includegraphics[width=0.7\linewidth]{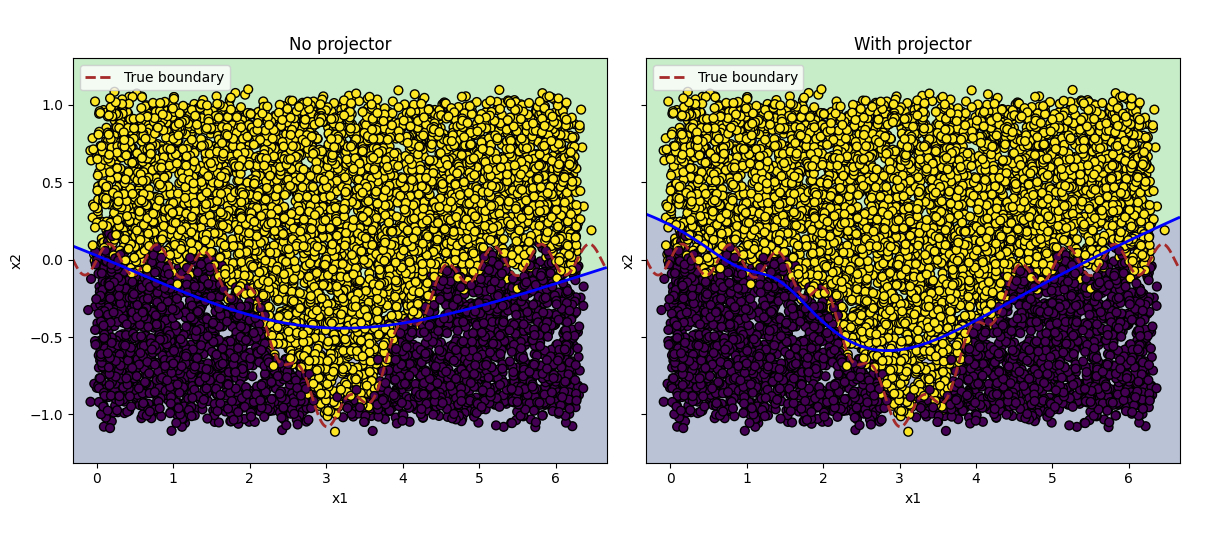}
\caption{Figure 6 compares the learned decision boundaries of the base model with and without the projector in test2 with 8000 and 2000 training and test samples accordingly after 15 epochs in Test 3.}
\end{figure}

\begin{figure}[H]
\centering
\includegraphics[width=0.7\linewidth]{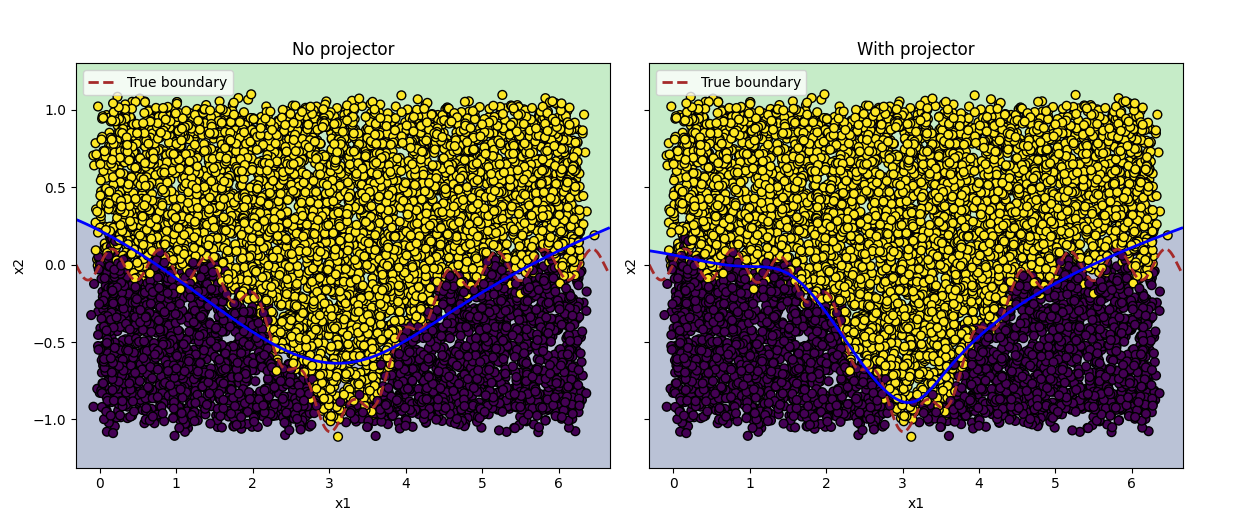}
\caption{Figure 7 compares the learned decision boundaries of the base model with and without the projector in test2 with 8000 and 2000 training and test samples accordingly after 40 epochs in Test3.}
\end{figure}

\begin{figure}[H]
\centering
\includegraphics[width=0.8\linewidth]{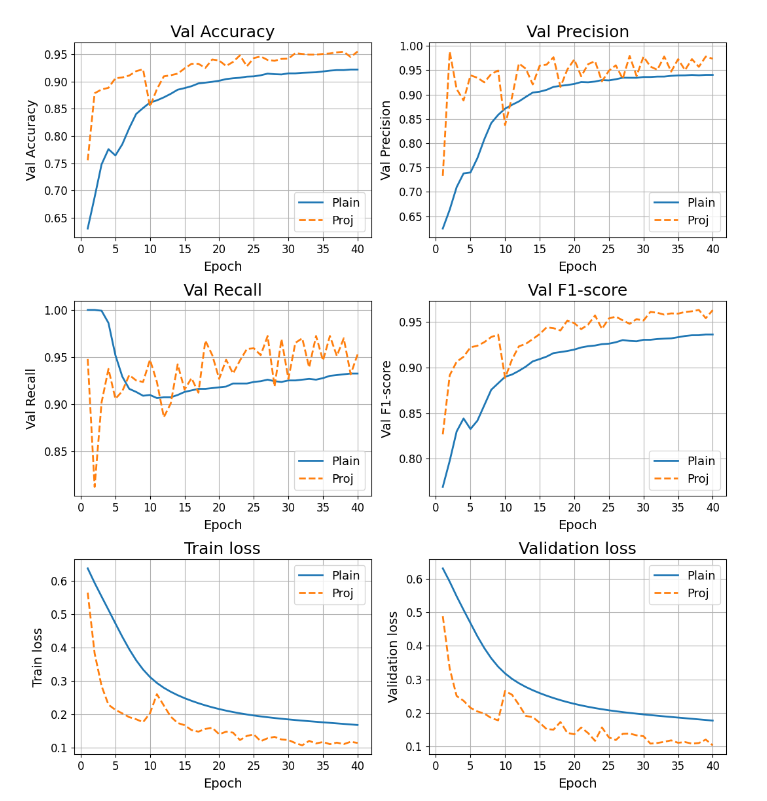}
\caption{Figure 8 Validation metrics and losses with training history in Test3.}
\end{figure}

Figure 7 compares the learned decision boundaries of the base model with and without the projector in Test 3, using 8,000 training samples and 2,000 test samples after 40 training epochs. With the increased number of epochs, the learned boundary obtained with the projector closely matches the global structure of the true decision boundary, whereas the model without the projector shows noticeably poorer global alignment. Figure 8 presents the validation metrics and losses over 40 training epochs.

With the higher number of samples additional improvement can be achieved if the projector applied iteratively more than 1 time:
Initialize \( h^{(0)} = h \). For \( k = 0, 1, \ldots, n - 1 \),
\[
h^{(k+1)} = \alpha h^{(k)} + (1 - \alpha) P\!\left(h^{(k)}\right).
\]
Figure 9 demonstrates improved fitting of the learned boundary using two projection steps relative to the previous case.

\begin{figure}[t]
\centering
\includegraphics[width=0.8\linewidth]{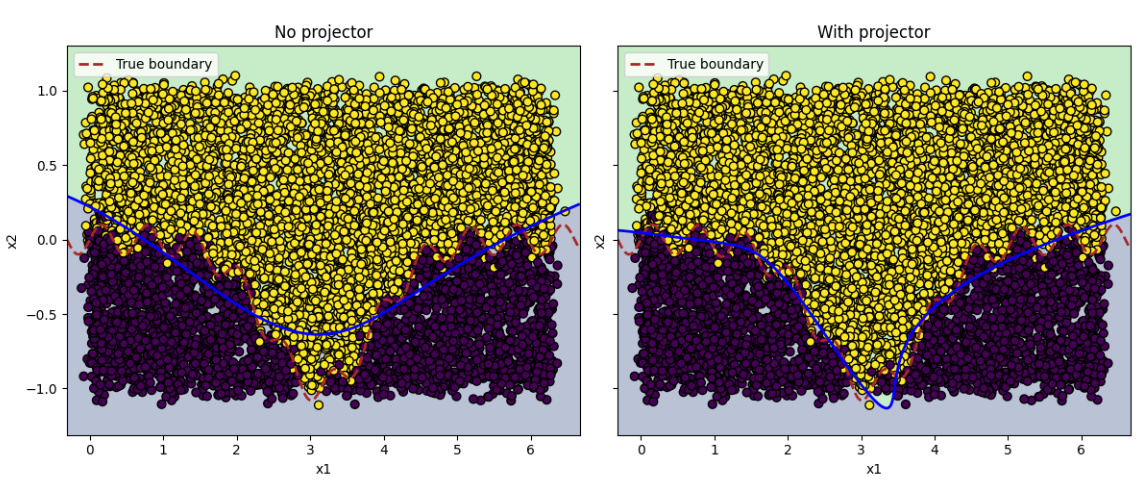}
\caption{Figure 9. Test 3 with 40 epochs and 2 projection steps. Additional smoothing by the 2nd projection steps made the learned boundary and original decision boundary almost coincide.}
\end{figure}

\subsection{Conclusions}
\begin{enumerate}
\item We can state that these results indicate that the proposed approach scales
well even when the number of samples is substantially increased with as much as 100 times more samples.
\item From validation metrics and losses history curves we can see that the
training dynamics observed with the projector is characterized by mild
non-monotonicity in both validation metrics and loss curves. This behavior can be
attributed to competing update components, where local gradient-driven updates
interact with global, projector-induced corrective steps. While these components
are not always perfectly aligned, their interaction enables stronger global shape
adaptation, resulting in faster convergence and improved alignment with the true
decision boundary compared to the plain model.
\end{enumerate}

\section{Transformer–based classifier with multi-scale convex projector}

As discussed earlier, the proposed enhancement can be integrated into components such as attention mechanisms, transformer layers, and feed-forward blocks that are standard elements of modern language models in form of
\[
h \leftarrow \alpha h + (1 - \alpha) P(h)
\]

\subsection{Dual Projector}
We introduce a \emph{Dual Projector} that performs multiscale smoothing in both
the feature and sequence (temporal) dimensions of the hidden representation
\( h \in \mathbb{R}^{B \times T \times D} \). Given the post-attention, post-MLP
state \( h \), the model applies an iterative convex combination
\[
h^{(k+1)} = \alpha_h h^{(k)} + (1 - \alpha_h) P\!\left(h^{(k)}\right),
\qquad \alpha_h \in (0,1),
\]
where \( P \) is the projector and
\( \alpha_h = \sigma(\operatorname{logit}_{\alpha_h}) \) is learned. The
projector can act on features, on sequence, or on both.

\paragraph{Feature projector (restriction--prolongation).}
For feature smoothing, we define a learnable restriction
\( Q^{*} : \mathbb{R}^{D} \to \mathbb{R}^{D_c} \) and prolongation
\( Q : \mathbb{R}^{D_c} \to \mathbb{R}^{D} \) (implemented as linear maps without
bias). Given \( x \in \mathbb{R}^{B \times T \times D} \), restriction produces
coarse features
\[
z = Q^{*} x \in \mathbb{R}^{B \times T \times D_c}.
\]

To ensure consistency between restriction and prolongation, we compute the
coarse coefficients by solving a small linear system
\[
A u = z, \qquad A = Q^{*} Q + \varepsilon I,
\]
(where \( \varepsilon > 0 \) stabilizes the solve), and then prolong back:
\[
\mathcal{P}_{\text{feat}}(x) = Q u .
\]

This realizes the oblique (and not strictly orthogonal) projector \( Q (Q^{*} Q)^{-1} Q^{*} \), yielding a
smoothed feature representation that preserves components representable on the
coarse feature subspace.
\subsection{Sequence projector (temporal restriction--prolongation)}
For temporal smoothing, we learn a low-rank temporal basis
\( W \in \mathbb{R}^{T \times T_c} \) and compute an orthonormal basis
\( Q_t \) via QR decomposition. The temporal projector is
\[
P_t = Q_t Q_t^{\top} \in \mathbb{R}^{T \times T},
\]
and acts as
\[
\mathcal{P}_{\text{seq}}(x) = P_t x,
\]
which restricts the sequence to a coarse temporal subspace and prolongs it back
without inversion.

In other words, our sequence projector is classical in the sense that it is symmetric, idempotent, and orthogonal, while remaining learnable because it is constructed from a trainable matrix W or temporal coarse subspace. The sequence dimension possesses an intrinsic geometric ordering, with natural coordinates such as time or token position, making orthogonal projection both natural and stable. In contrast, the feature dimension represents a learned latent space without a canonical geometry; therefore, we adopt a more flexible oblique projector to avoid over-constraining representation learning.

\subsection{Dual action.}
When both modes are enabled, the projector composes temporal and feature actions:
\[
\mathcal{P}(x) = \mathcal{P}_{\text{feat}}\!\left(\mathcal{P}_{\text{seq}}(x)\right),
\]
providing simultaneous smoothing across time and features. Repeating the update
for a small number of steps (e.g., 2--3) yields a controllable coarse-to-fine
correction that improves stability and generalization while remaining fully
differentiable and lightweight.

\subsection{Multi-Scale Convex Projector}

We introduced a dual pseudo-projector that performs smoothing in both the feature and sequence representation spaces. Now, we will generalize this approach to multiple coarse levels, but we construct the multi-scale hierarchy exclusively along the feature dimension. This enables theoretically grounded coarse-to-fine corrections that act as feature-space smoothers during optimization. In contrast, the sequence dimension is ordered and sample-dependent; coarsening operations along this axis (e.g., pooling or subsampling) are inherently nonlinear and do not admit a convex projector interpretation. For this reason, multi-scale structure is restricted to the feature dimension, while sequence-level variability is handled by architectural mechanisms.
Different samples, layers, and training stages benefit from different levels of abstraction. Using multiple coarse sizes lets the model adaptively balance stability vs. expressiveness, capturing both global structure and mid-scale patterns without locking the optimization into a single bias. We introduce a multi-scale convex projector that combines several projectors operating at different coarse dimensions. Let
\[
P_i = Q_i \left( Q_i^{\top} Q_i \right)^{-1} Q_i^{\top},
\qquad i = 1, \ldots, K .
\]
denote a family of linear projectors parameterized by learnable
restriction--prolongation operators \( Q_i^{\top} \) and \( Q_i \), each
corresponding to a distinct coarse dimension \( D_{c,i} \).

We define the multi-scale projector as a convex combination

\begin{equation}
P_{\mathrm{MS}} = \sum_{i=1}^{K} \alpha_i P_i,
\qquad
\alpha_i \ge 0,
\qquad
\sum_i \alpha_i = 1,
\end{equation} 

where the coefficients \( \alpha_i \) are learned jointly with the model
parameters. In practice, we parameterize the coefficients via unconstrained
logits \( \beta_i \) and enforce the simplex constraint using a softmax,
\[
\alpha_i = \frac{\exp(\beta_i)}{\sum_j \exp(\beta_j)} .
\]
Feature representations lie in a linear vector space where coarse abstractions can be defined via linear projections. Individual projectors are designed to be approximately idempotent and non-expansive. While the convex combination of projectors with learned restriction and prolongation is not strictly idempotent, the use of normalized non-negative weights preserves non-expansiveness under standard operator norm assumptions and enables controlled coarse-scale interpolation without amplification.
\emph{This allows the model to adaptively emphasize different coarse resolutions
during training}. Although \( P_{\mathrm{MS}} \) is not itself an idempotent
projector, it acts as a learnable smoothing operator that interpolates between
multiple coarse scales.

The projector is applied through a residual correction
\[
h \leftarrow h + \eta \bigl( P_{\mathrm{MS}} h - h \bigr),
\]
where \( \eta \in (0,1) \) controls the strength of the coarse-scale correction.
The parameter \( \eta \) may be treated in one of two ways:
\begin{itemize}
    \item as a learnable scalar, optimized jointly with the network parameters; or
    \item as a deterministically scheduled parameter (e.g., increasing or decreasing over training epochs) to regulate the influence of the projector throughout training.
\end{itemize}
This flexibility allows coarse smoothing to be
emphasized during early training or adaptively tuned based on the task.

\subsection{Test Model}
As a test model we propose a weight-tied transformer encoder that performs iterative refinement of token representations. Instead of increasing depth, the same attention–MLP block is repeatedly applied (we tested 2 times case) to approximate a fixed-point representation, with an optional coarse-scale projector used as a post-refinement correction. Our purpose was to compare training dynamics with / without multi-scale projector. 
A simplified model workflow representation is given in Figure 10 
\begin{figure}[H]
\centering
\includegraphics[width=0.8\linewidth]{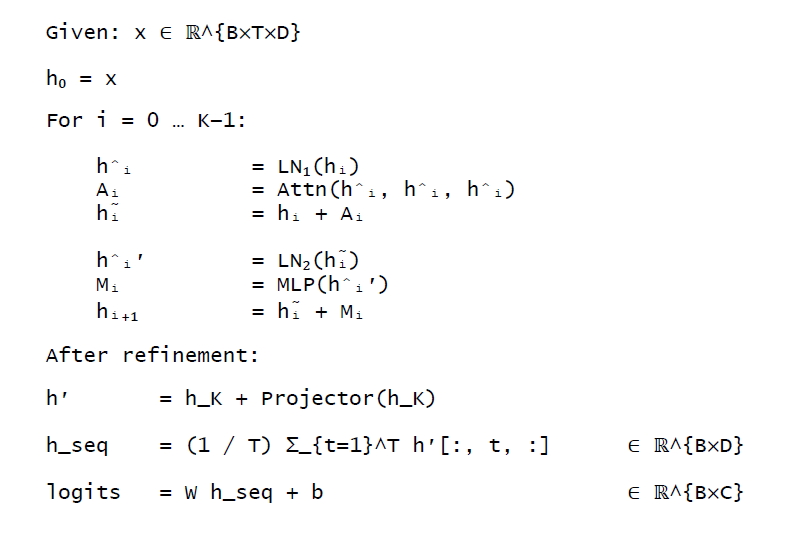}
\caption{Figure 10. Simplified model workflow}
\end{figure}

As we experimented with various text samples, some of which contain a large
number of tokens, we used the \texttt{allenai/longformer-base-4096} model as the
text encoder, which supports sequences of up to 4{,}096 tokens per sample. This
model is available at the Hugging Face hub
\textit{\href{https://huggingface.co/allenai/longformer-base-4096}{allenai/longformer-base-4096}} with an Apache-2.0
license.

\section{Numerical Study}

\subsection{Experiment with QQP dataset}

Quora Question Pairs (QQP) is a large-scale binary classification dataset of question pairs labeled for semantic equivalence, commonly used to evaluate paraphrase detection and sentence-pair similarity models, originally released by Quora and popularized through [19]. 
There are multiple valid approaches to dataset preparation and hyperparameter selection for this task. In this work, we adopt one representative experimental setup. We randomly sampled 30,000 examples and applied length-based filtering to ensure manageable input sequences. This preprocessing, together with enforcing class balance, resulted in a modest reduction in the final dataset size.

We evaluate three progressively more complex training configurations. In all cases the following settings was applied:
\\
\\
\begin{tabular}{l c}
\hline
\textbf{Parameter} & \textbf{Value} \\
\hline
Maximum number of tokens in the sample & 128 \\
Model training batch size & 32 \\
Embedding vector size & 768 \\
Feature coarse level sizes of projector & [16, 64, 128] \\
Sequence coarse level size of projector & 32 \\
Number of training epochs & 30 \\
Learning rate bounds (linear schedule) & [0.01, 0.0001] \\
Number of model refinement steps & 2 \\
Train / Test split & 80\% / 20\% \\
Parameter \( \eta \) (scheduled) &
\( 1 - 0.8 \cdot \text{epoch\_number} / \text{max\_epochs} \) \\
\hline
\end{tabular}
\\\\\\
We chose a manual schedule for parameter \( \eta \) that controls the contribution
of the projector's smoothing correction. Per our schedule, it provides the
maximal contribution at the beginning and linearly decays toward the end of
training. The intuition is that at the beginning the projector can help plunge
to the global minimum by ignoring high-frequency noise and later switch to more
local refinements.

As opposed to the selection of \( \eta \), the parameters \( \alpha_i \), which
control the contribution of individual projectors \( P_i \) corresponding to a
certain level of smoothing, are trainable. They are distributed according to a
softmax function that assigns convex weights serving as probabilities and keeps
their sum normalized to 1. In terms of PyTorch object definitions, it looks like this:
\\\\
\textit{
\(\alpha = \texttt{torch.softmax}(\texttt{self.alpha\_logits}, \texttt{dim}=0)\),
\\
where 
\\
\(\texttt{self.alpha\_logits} = \texttt{nn.Parameter}(\texttt{torch.zeros}(\texttt{len}(\texttt{Dc\_list})))\),
\\
and
\\
\(\texttt{Dc\_list}\) is the list of all coarse subspace sizes.
}
\subsubsection{1st case: Balanced data - 50\% negative / 50\% positive. No noise added to data samples}
We compare performance of 2 models – Plain (no Projector) and Proj (with Projector)
Figure 11a and 11b demonstrate the training performance. In each epoch four evaluation metrics, train and validation loss and gradient norm behavior are depicted. Visually performance of these two models is quite close. There are multiple ways to compare model performance numerically. One criterion for selecting the best model is to compute the average value of evaluation metrics—accuracy, precision, recall, and F1 score—across training epochs. 
\[
\text{Average} \approx \frac{1}{T} \sum_{e=1}^{T} \text{metric}(e),
\]
where \(\text{metric}(e)\) denotes the value of a given evaluation metric
(e.g., accuracy, precision, recall, or F1 score) at epoch \( e \), and \( T \)
is the total number of training epochs. Figure 12 shows that the Proj model in this comparison is a winner in terms of all four metrics.
\begin{figure}[t]
\centering
\includegraphics[width=0.7\linewidth]{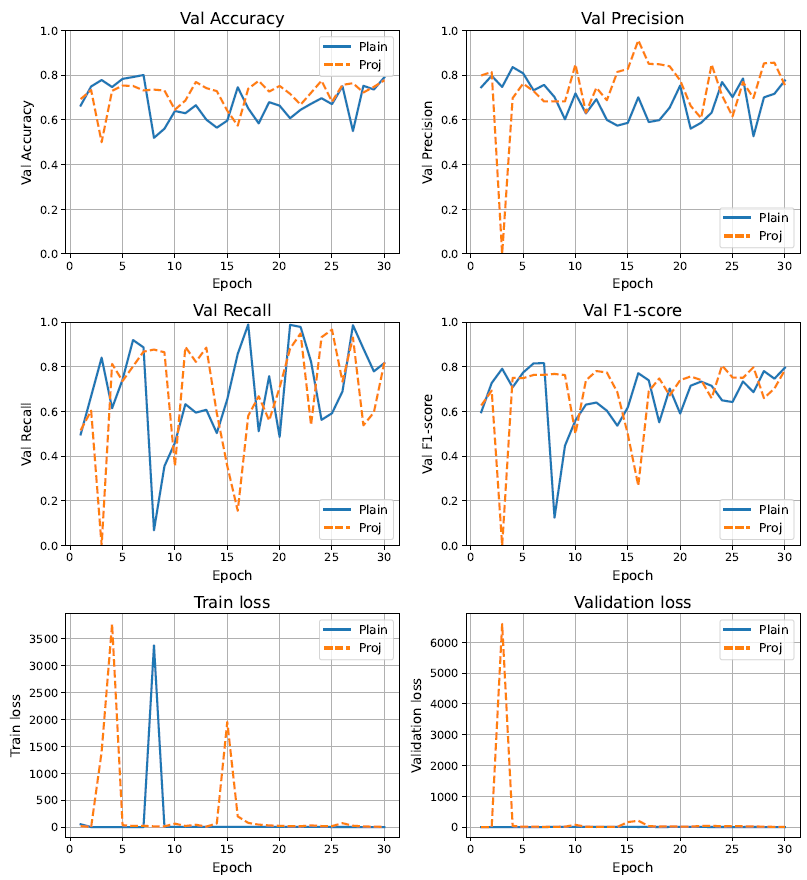}
\caption{Figure 11a. QQP dataset, Case 1 training / evaluation history }
\end{figure}

\begin{figure}[t]
\centering
\includegraphics[width=0.5\linewidth]{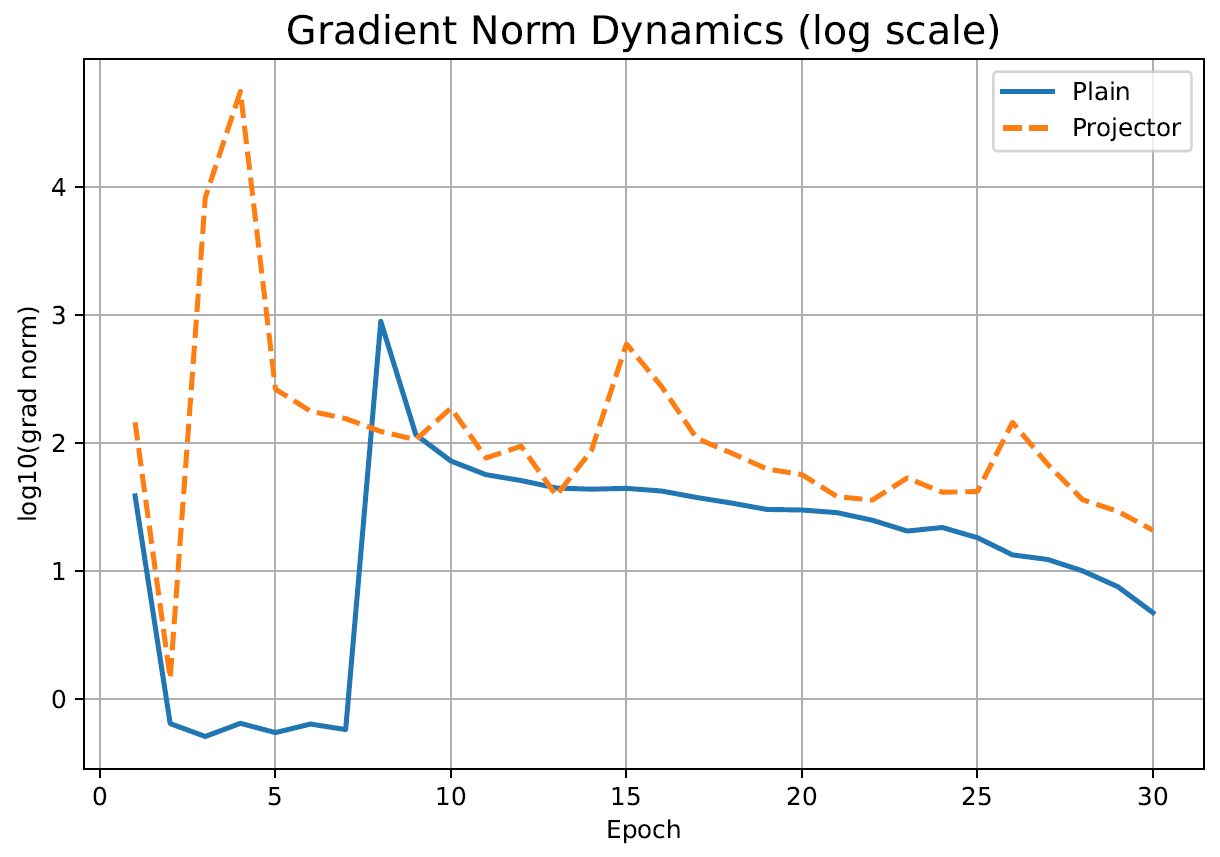}
\caption{Figure 11b. QQP dataset, Case 1. Gradient norms.}
\end{figure}

\begin{figure}[t]
\centering
\includegraphics[width=0.8\linewidth]{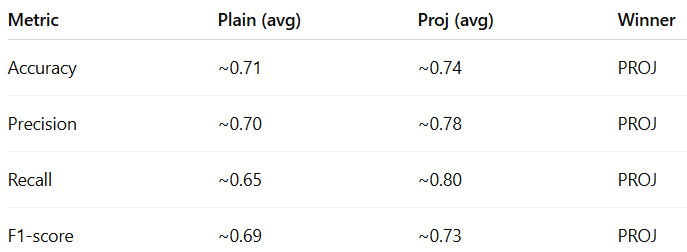}
\caption{Figure 12. QQP dataset, Case 1. When averaged across the full training trajectory, the Projector model consistently outperforms the Plain baseline in accuracy, precision, recall, and F1-score, indicating superior expected performance despite higher variance.}
\end{figure}

\subsubsection{2nd case: imbalanced data - 70\% negative / 30\% positive. No noise added to data samples}

In this case, we challenge the models by violating class balance. Training on imbalanced data is known to bias optimization toward majority classes, often leading to misleadingly high accuracy while degrading recall and F1 score for minority classes. This setting provides a more stringent evaluation and can highlight the advantages of the projector-based model, which encourages global consistency and structured representations that are less sensitive to class imbalance.
Figure 13a and 13b demonstrate the training performance. In each epoch four evaluation metrics, train and validation loss and gradient norm behavior are depicted. As was mentioned, when the dataset has a moderate to strong imbalance, F1-score becomes a more important measure than accuracy, that can be biased to the dominant class.
From Recall and F1-score values, we may conclude that the Plain version may not train well enough and Figure 14 shows that the Proj model is a winner in terms average metrics.

\begin{figure}[H]
\centering
\includegraphics[width=0.7\linewidth]{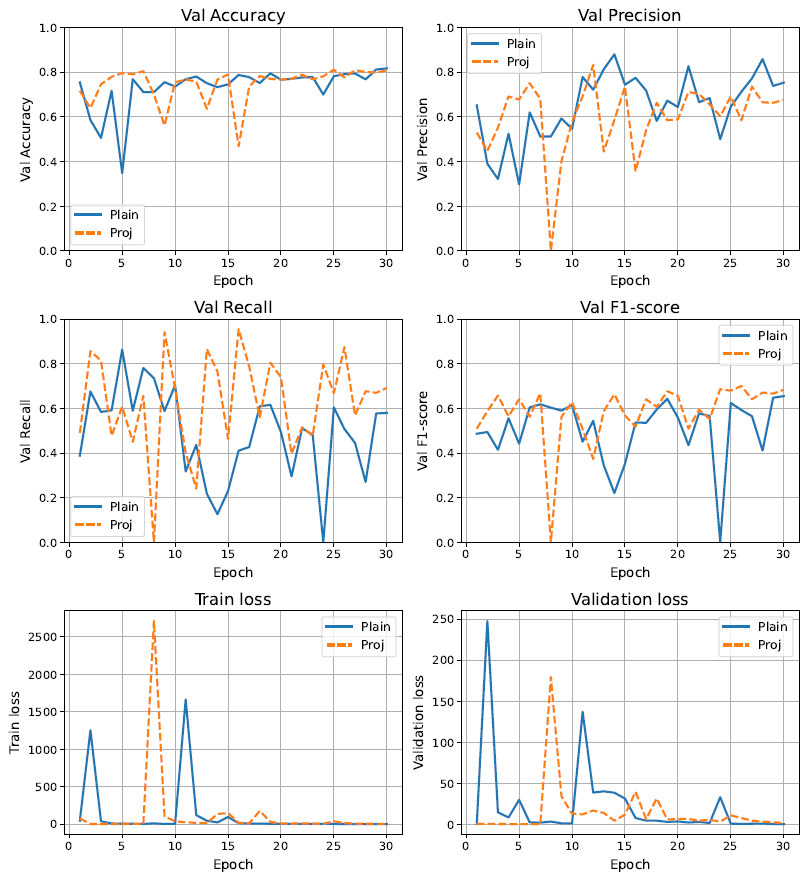}
\caption{Figure 13a. QQP dataset, Case 2 training / evaluation history}
\end{figure}

\begin{figure}[H]
\centering
\includegraphics[width=0.5\linewidth]{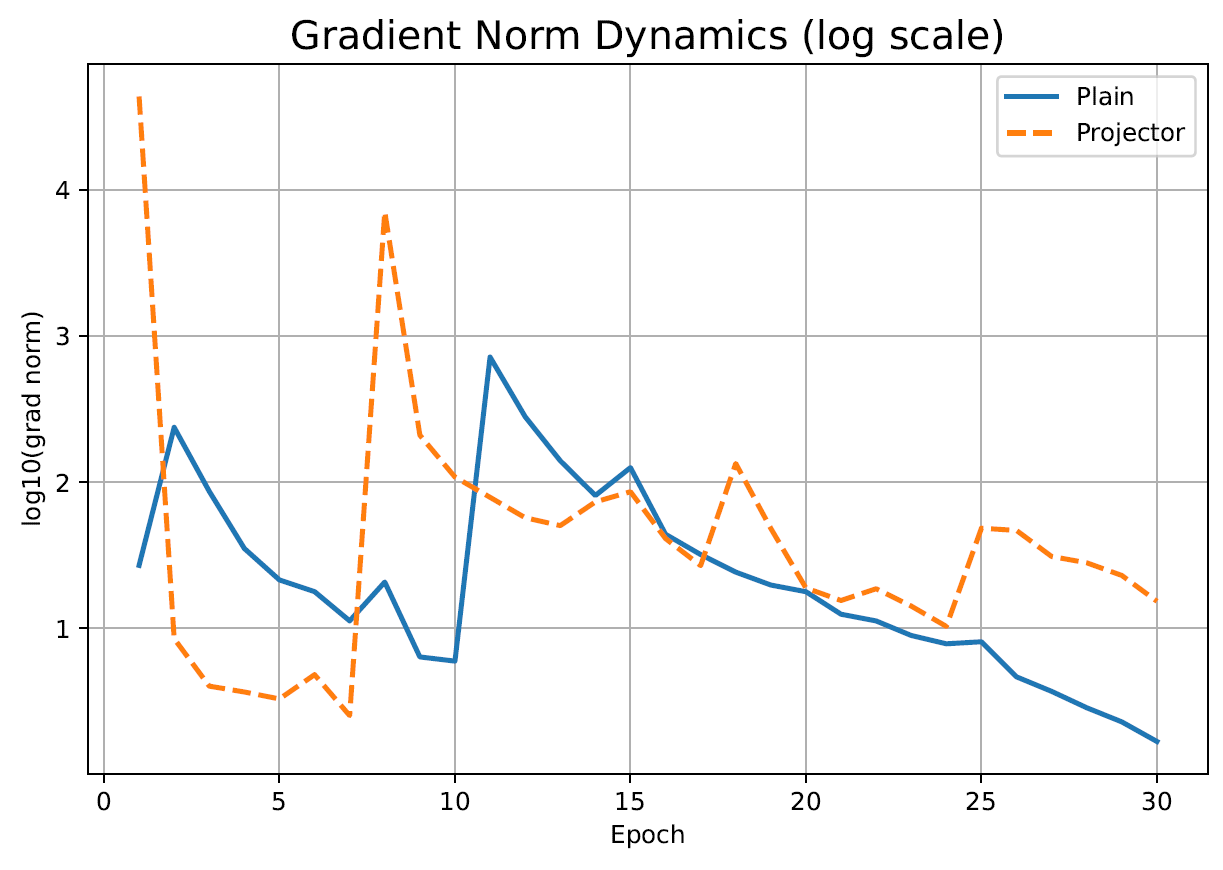}
\caption{Figure 13b. QQP dataset, Case 2. Gradient norms.}
\end{figure}

\begin{figure}[H]
\centering
\includegraphics[width=0.7\linewidth]{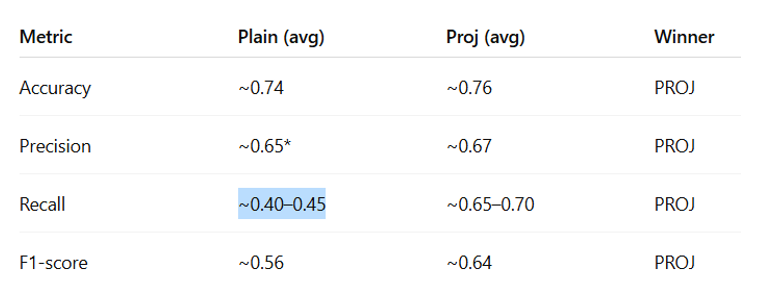}
\caption{Figure 14. QQP dataset, Case 2. When averaged across the full training trajectory, the Projector model consistently outperforms the Plain baseline in accuracy, precision, recall, and F1-score, indicating superior expected performance despite higher variance.}
\end{figure}
\noindent Figure 15 provides a point-wise comparison showing positive differences of Proj and Plain accuracies and F1-scores.
\begin{figure}[H]
\centering
\includegraphics[width=0.6\linewidth]{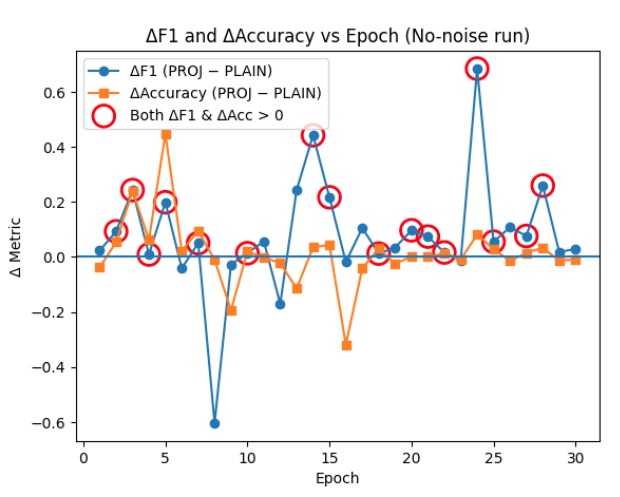}
\caption{Figure 15. Epoch-wise differences in Accuracy and F1 score between Proj and Plain models.}
\end{figure}

\subsubsection{3rd case: imbalanced data - 70\% negative / 30\% positive. Noise added to data samples}

Every sample in this dataset consists of Premise and Hypothesis and labeled positive if they semantically close. As a next level to challenge the model we augment the premise text by injecting sentence-level noise. Noise is added probabilistically and consists of randomly selected, semantically unrelated sentences drawn from a fixed noise pool. For each training or evaluation example, noise is applied independently as follows. With probability p=0.7, the original premise is augmented; otherwise, it is left unchanged. When augmentation is triggered, an integer k is sampled uniformly from {1,…,K}, where K=10 is the maximum number of noise sentences. The selected k noise sentences are sampled without replacement from a predefined list of noise sentences. To avoid positional bias, the order of all sentences (the original premise and the injected noise sentences) is randomly shuffled before concatenation. 
\\\\
\textbf{Intuition behind Noise Injection} is that
injecting sentence-level noise deliberately increases the number of 
\emph{label-irrelevant directions} in the input representation space. 
These directions contribute to variance but do not correlate with the target signal.
In a plain model, all token or sentence embeddings are aggregated within the same feature space. 
Because there is no explicit mechanism to separate relevant from irrelevant components, 
noise contributes additively to the final representation leading to \emph{increased variance and decreased signal-to-noise ratio}.
This effect becomes especially pronounced when the injected noise dominates 
in length.
\emph{A projection-based model explicitly constrains representations to a lower-dimensional subspace aligned with a task-relevant signal while suppressing irrelevant directions introduced by noise}.
\\\\
Let’s follow the same evaluation procedure as before to compare the two models. Figure 16 clearly demonstrates advantage of the Proj model over Plain model in terms of validation metrics history

\begin{figure}[H]
\centering
\includegraphics[width=0.7\linewidth]{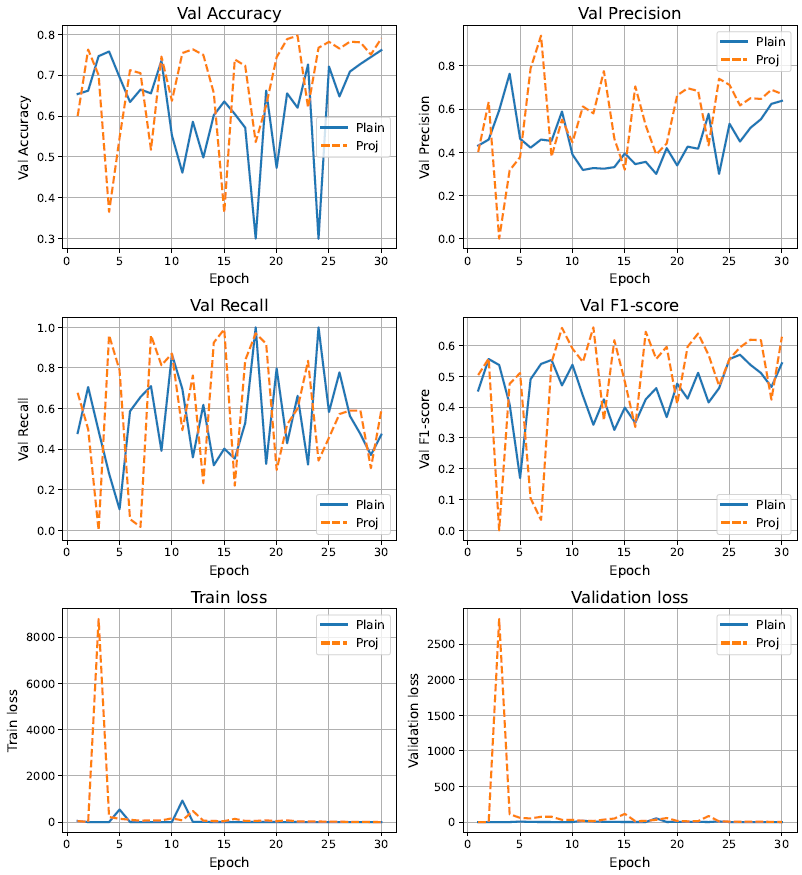}
\caption{Figure 16. QQP dataset, Case 3 training / evaluation history}
\end{figure}
Averaged metrics table in Figure 17 confirms that Proj model substantially outperforms the Plain model. In fact, the Plain model failed to train according to the metrics. This statement is supported by the “Best Epochs” table given in Figure 18. 
\begin{figure}[H]
\centering
\includegraphics[width=0.8\linewidth]{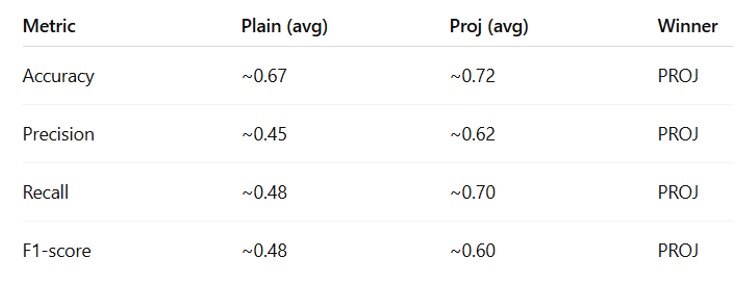}
\caption{Figure 17. QQP dataset, Case 3. Averaged validation metrics across the full training trajectory.}
\end{figure}

\begin{figure}[H]
\centering
\includegraphics[width=0.8\linewidth]{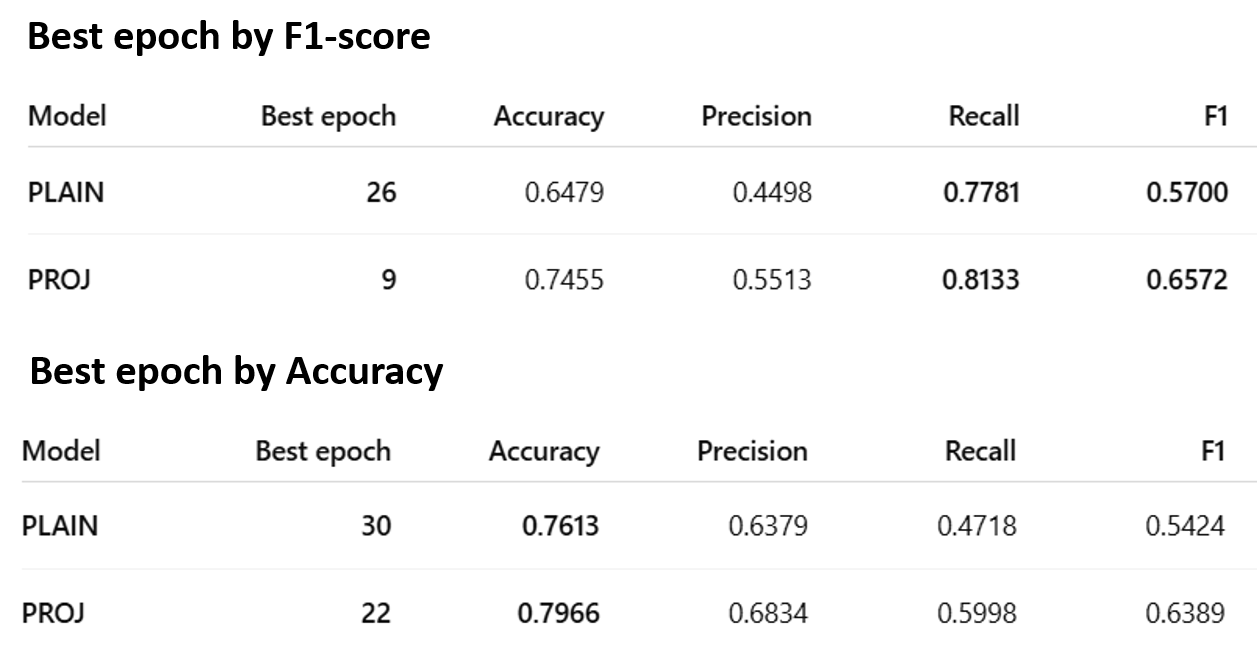}
\caption{Figure 18. QQP dataset, Case 3. “Best Epoch” in terms of Accuracy and F1-score in training competing models}
\end{figure}

\subsection{Experiment with SNLI dataset}

The SNLI (Stanford Natural Language Inference) dataset is a large-scale benchmark for natural language inference. It consists of sentence pairs—a premise and a hypothesis—annotated with one of three labels: entailment, contradiction, or neutral. The premises are derived from image captions, while the hypotheses are human-written to reflect the three inference relations, making SNLI a standard dataset for training and evaluating sentence understanding and reasoning models. We used only entailment and contradiction labeled samples to keep classification binary.  

Using this dataset, we ran similar experiments and observed the same trends in
metrics and losses across epochs. We further challenged the models by increasing
class imbalance to an 80--20\% split. As emphasized earlier, we expected the
projector’s smoothing and global effects to positively address this imbalance.
The intuition is that projection-based smoothing discourages reliance on frequent
but weakly informative features that dominate majority-class samples, while
amplifying minority-class signals by reducing interference from
majority-dominated feature directions.

After filtering and imposing the imbalance restriction, we obtained a random
set of 21{,}468 samples, with 17{,}174 training samples and 4{,}294 test samples,
both containing 20\% positive cases. The only change in the training parameters
was setting the lower bound of the learning rate to \(5 \times 10^{-3}\),
compared to \(1 \times 10^{-2}\) in the previous experiment, resulting in more
conservative parameter updates during later training stages.

Figure 19 shows a clear advantage of the Proj model over the Plain model, particularly in terms of the F1 score, which serves as a precision–recall aggregator and is especially informative for imbalanced datasets. In such settings, accuracy can remain high due to bias toward the majority class in low-prevalence data, while precision and recall often degrade substantially.
\begin{figure}[H]
\centering
\includegraphics[width=0.8\linewidth]{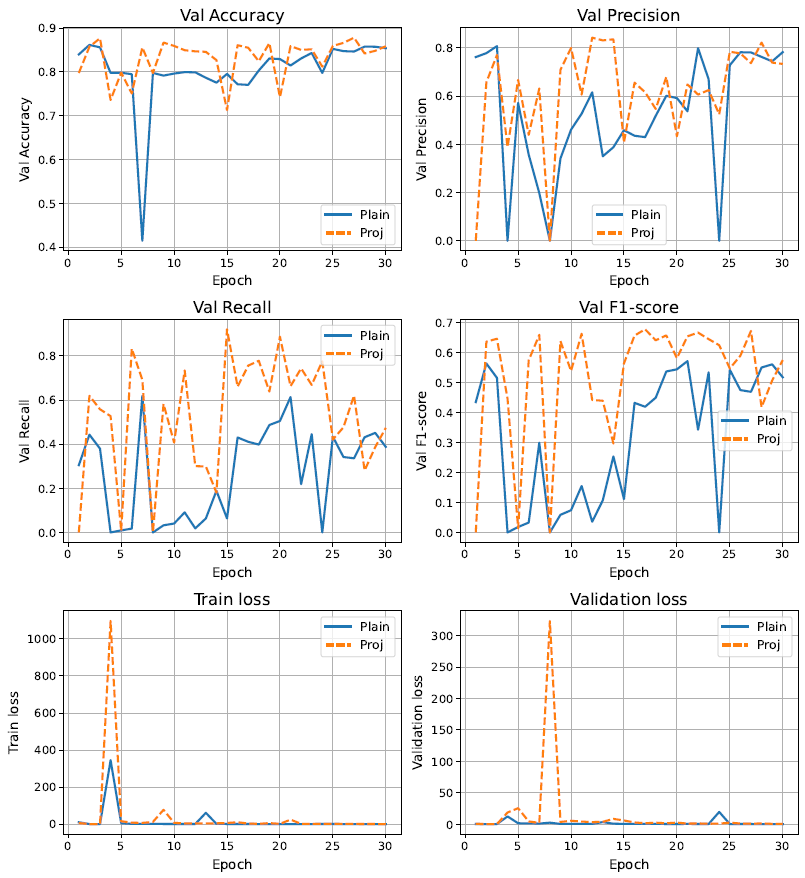}
\caption{Figure 19. SNLI dataset, training / evaluation metrics history}
\end{figure}

Figure 20 demonstrates the gradient norms’ behavior. From a multigrid perspective, the elevated gradient norms observed for the Proj model in the early epochs correspond to coarse-level corrective steps that act globally on the representation space. The projector induces strong early updates that reshape the global structure of the decision boundary, leading to higher gradient magnitudes before convergence stabilizes.
\begin{figure}[H]
\centering
\includegraphics[width=0.5\linewidth]{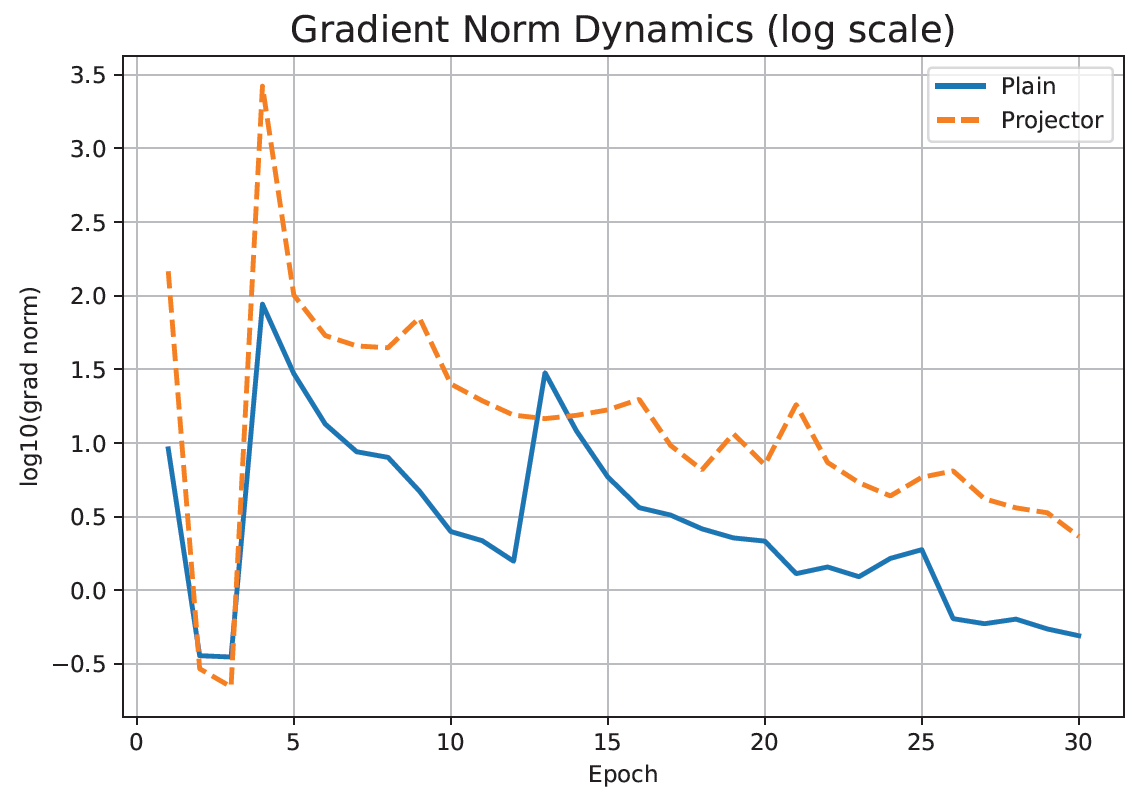}
\caption{Figure 20. The much higher (this is a log graph) early gradient norms in the Proj model are consistent with a multigrid-like coarse correction phase, where global error components are addressed first, followed by fine-scale refinement as training converges.}
\end{figure}

\noindent
Figure 21 shows “the best epochs” for both models. It demonstrates that he projector-corrected model reaches its maximum performance metrics at an early stage, whereas the plain model never achieves acceptable recall. 
\begin{figure}[H]
\centering
\includegraphics[width=0.8\linewidth]{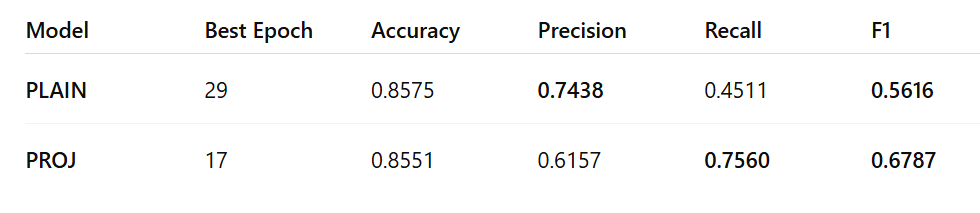}
\caption{Figure 21 shows that the projector-corrected model reaches its maximum performance metrics at an early stage, whereas the plain model never achieves acceptable recall.}
\end{figure}
\noindent
As mentioned earlier, in all practical examples the subspace dimensions of the projectors in (3) are [16, 64, 128], respectively. The parameters in (3) are learnable and allow the model to automatically determine the contribution of each projector component. Figure 22 shows behavior of the weights \(\alpha_0\),  \(\alpha_1\), \(\alpha_2\) over training epochs. As we can see, all three components make a meaningful contribution; however, the largest contribution comes from the subspace with the lowest dimension. This observation may help guide adjustments to the subspace sizes, potentially improving training effectiveness.

\begin{figure}[H]
\centering
\includegraphics[width=0.6\linewidth]{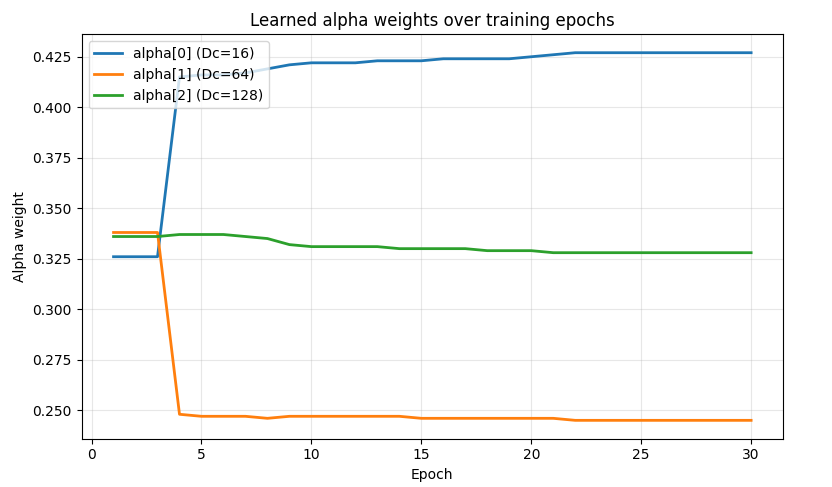}
\caption{Figure 22. Behavior of the weights \(\alpha_0\),  \(\alpha_1\), \(\alpha_2\) over training epochs.}
\end{figure}

\subsection{Experiments with MIMIC-IV discharge summaries}
The MIMIC-IV discharge summary dataset available through physionet.org consists of long unstructured clinical discharge notes documenting patient hospital stays. These texts are characterized by extreme length, heterogeneous structure, and a high density of medically irrelevant or redundant information relative to downstream prediction targets.

The classification task was to predict whether a patient would be readmitted within 30 days after discharge. For evaluation, we retrieved a set of 17,000 random samples with a balanced class distribution (50/50). To ensure compatibility with our tokenizer, we imposed a maximum sequence length of 4,096 tokens. We selected projector subspace sizes of [16, 64, 128], a batch size of 16 to fit the available GPU memory, and scheduled the learning rate from 
\(5 \times 10^{-3}\) to \(1 \times 10^{-4}\).

Prediction based on very long, unstructured, and noisy clinical records is known to be a challenging task. From our experience, validation metrics typically fall within the 55\%--70\% range. However, the objective of this study is not to achieve the highest possible predictive accuracy, but rather to compare the relative performance of two models.

Figure 23 shows the validation metrics across training epochs, as was done in the previous examples. As can be seen, with the specified set of coarse subspaces, reasonably good validation metrics for this task are achieved \emph{within just one epoch}. These results are marked with orange circles. The same metrics at the final epoch (epoch 30) are marked with green circles, showing that after 30 epochs the training did not achieve better performance than that obtained after the first epoch.

Intuitively, this effect can be explained by the fact that the multi-scale projector produced from the very beginning a gradient that more directly targets the global optimum, reducing the need for many incremental steps that might otherwise become trapped in local optima.

\begin{figure}[H]
\centering
\includegraphics[width=0.7\linewidth]{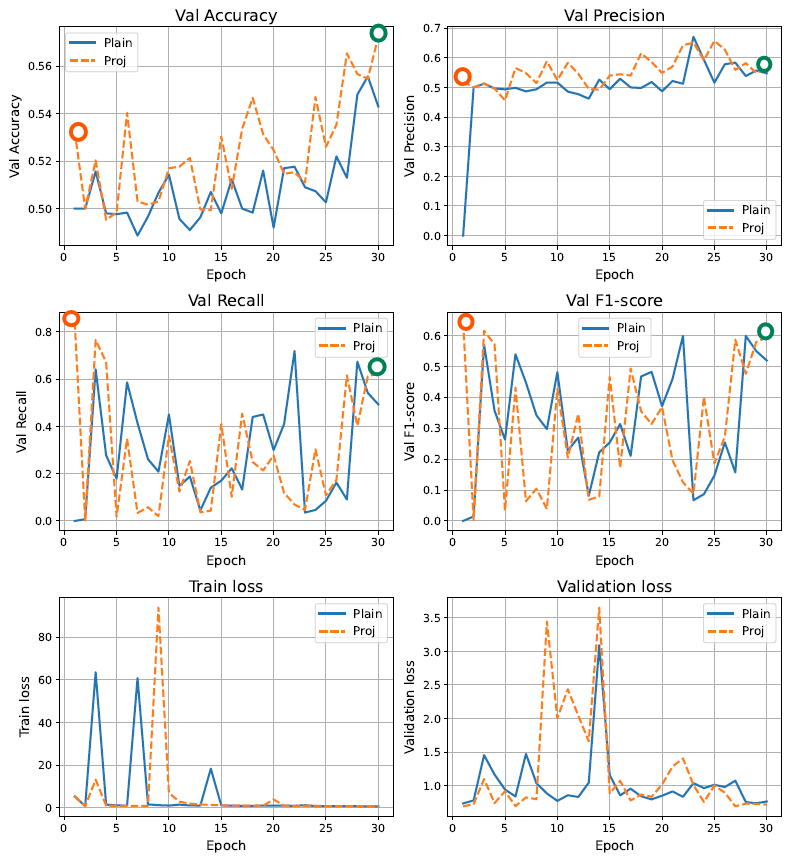}
\caption{Figure 23. MIMIC-IV discharge summaries dataset, training / evaluation metrics history}
\end{figure}

\section{Results and Discussion}

Across all experiments, the projector-augmented (Proj) models consistently
exhibit improved training dynamics and generalization behavior compared to the
plain baseline. These improvements are observed in both synthetic and real-world
settings and become increasingly pronounced as task difficulty increases,
including non-convex decision boundaries, class imbalance, and label-irrelevant
noise.

A recurring pattern is that the Proj model converges more rapidly to solutions
with better global structure. In low-dimensional synthetic experiments, this is
directly visible in the learned decision boundaries, where projection-based
smoothing enables stronger alignment with the true global shape while reducing
sensitivity to high-frequency distortions. In higher-dimensional text
classification tasks, the same effect manifests through faster stabilization of
validation metrics and improved recall and F1 score, particularly in settings
where accuracy alone is misleading.

The advantage of the projector becomes especially clear under class imbalance.
While the plain model often achieves high accuracy by favoring the majority
class, it suffers from poor recall and F1 score for minority classes. In
contrast, the Proj model maintains substantially better precision--recall
balance. This suggests that projection-based smoothing discourages reliance on
frequent but weakly informative features and mitigates gradient dominance from
majority-class samples, thereby amplifying minority-class signal.

Gradient norm analysis provides further insight into the underlying optimization
behavior. The Proj model exhibits elevated gradient norms during early training
epochs, followed by gradual stabilization. From a multigrid perspective, this is
consistent with an initial coarse correction phase that addresses global error
components before transitioning to finer-scale refinement. The plain model, by
comparison, shows more rapid gradient decay, often converging prematurely to
majority-biased or locally suboptimal solutions.

The benefits of the projector are further amplified in the presence of
sentence-level noise. Injecting semantically irrelevant content increases
variance in the hidden representations and obscures task-relevant structure for
the plain model, frequently preventing successful training. The Proj model
remains robust in this regime, indicating that the learned projection operator
effectively suppresses label-irrelevant directions while preserving discriminative
information.

Overall, these results demonstrate that the proposed pseudo-projector acts as an
implicit regularizer that improves convergence speed, stability, and
generalization without modifying the loss function, optimizer, or core model
architecture. The effect is most pronounced in challenging regimes where global
structure must be recovered from noisy, imbalanced, or highly non-convex data,
while no adverse effects are observed in simpler or balanced settings.

\section{Conclusion}

We introduced an approach that is a lightweight modification which can be integrated into existing language models and other neural networks without altering their core architecture. The core component of this approach is a pseudo-projector that exploits smoothing functionality to suppress irrelevant components of the representation. We demonstrate its effectiveness as a lightweight enhancement for neural network training. By operating directly on hidden representations, the proposed method
acts as a corrective mechanism that improves convergence behavior, stability,
and generalization without modifying the loss function or the core architecture
of the model.

Our results suggest that the pseudo-projector can be naturally used as a
representation-level corrector for transformer-based models, where it may be
applied after attention or feed-forward blocks to enforce global consistency and
suppress label-irrelevant variability. Across synthetic and real-world text
classification tasks, the projector consistently improves performance,
particularly under challenging conditions such as class imbalance, input noise,
and highly non-convex decision boundaries.

This work represents an ongoing research toward incorporating multigrid-inspired
corrective operators into modern neural architectures. It aims to
extend this approach to large-scale language models, investigate adaptive
scheduling and placement strategies, and further analyze the theoretical
properties of projection-based smoothing in high-dimensional representation
spaces.

The corrective projector described in this study is applicable to a broader class of models beyond transformer-based or language-oriented architectures. However, our primary focus in this work is on language-related applications, motivated by ongoing projects in unstructured and noisy medical notes processing and prediction.

\section*{References}

\begin{enumerate}

\item A.~A.~Fedorenko,
\emph{The speed of convergence of one iterative process},
USSR Computational Mathematics and Mathematical Physics,
vol.~4, no.~3, pp.~227--235, 1964.

\item A.~Brandt,
\emph{Multi-level adaptive solutions to boundary-value problems},
Mathematics of Computation,
vol.~31, no.~138, pp.~333--390, 1977.

\item J.~W.~Ruge and K.~St{\"u}ben,
\emph{Algebraic multigrid},
in \emph{Multigrid Methods},
Frontiers in Applied Mathematics, SIAM, 1987.

\item W.~Hackbusch,
\emph{Multi-Grid Methods and Applications},
Springer Series in Computational Mathematics,
Springer, 1985.

\item A.~Brandt and O.~E.~Livne,
\emph{Multigrid Techniques: 1984 Guide with Applications to Fluid Dynamics},
SIAM, 1984.

\item V.~Bulgakov,
\emph{Multi-level iterative technique and aggregation concept with semi-analytical preconditioning for solving boundary-value problems},
Communications in Numerical Methods in Engineering,
vol.~9, pp.~649--657, 1993.

\item V.~Bulgakov and G.~Kuhn,
\emph{High-performance multilevel iterative aggregation solver for large finite-element structural analysis problems},
International Journal for Numerical Methods in Engineering,
vol.~38, pp.~3529--3544, 1995.

\item C.~W.~Oosterlee and T.~Washio,
\emph{Multigrid methods for optimization problems},
Journal of Computational and Applied Mathematics,
2000.

\item S.~G.~Nash,
\emph{A multigrid approach to discretized optimization problems},
Optimization Methods and Software,
vol.~14, no.~1--2, pp.~99--121, 2000.

\item J.~G{\"u}nther, L.~Ruthotto, B.~Schr{\"o}der, E.~Cyr, and N.~R.~Gauger,
\emph{Layer-parallel training of deep residual neural networks},
SIAM Journal on Scientific Computing,
2020.

\item Z.~Moon and E.~Cyr,
\emph{Parallel training of GRU networks with a multigrid solver for long sequences},
International Conference on Learning Representations (ICLR), 2022.\\
Available at: \url{https://arxiv.org/pdf/2203.04738}

\item E.~Cyr et al.,
\emph{TorchBraid: High-performance layer-parallel training of deep neural networks with MPI and GPU acceleration},
ACM, 2025.

\item T.-W.~Ke et al.,
\emph{Multigrid neural architectures},
2017.\\
Available at: \url{https://arxiv.org/pdf/1611.07661v2}

\item M.~Eliasof, J.~Ephrath, L.~Ruthotto, and E.~Treister,
\emph{MGIC: Multigrid-in-channels neural network architectures},
2022.\\
Available at: \url{https://arxiv.org/pdf/2011.09128}

\item J.~He and J.~Xu,
\emph{MgNet: A unified framework of multigrid and convolutional neural network},
Science China Mathematics,
2019.

\item Z.~Li et al.,
\emph{M2NO: An efficient multi-resolution operator framework for dynamic multi-scale PDE solvers},
2025.\\
Available at: \url{https://arxiv.org/pdf/2406.04822}

\item I.~Luz et al.,
\emph{Learning Algebraic Multigrid Using Graph Neural Networks},
2020.\\
Available at: \url{https://arxiv.org/pdf/2003.05744}

\item W. ~Rudin, 
\emph{Functional Analysis, 2nd ed., McGraw–Hill}, 
1991.\\

\item A.~Wang et al.,
\emph{GLUE: A multi-task benchmark and analysis platform for natural language understanding},
2018.
\end{enumerate}

\noindent
\textit{
Computations were performed using a single NVIDIA RTX 6000 GPU with 48 GB at Division
of Endocrinology, Mass General Brigham, Boston, MA
}

\end{document}